\documentclass[authoryear,preprint,5p,twocolumn]{elsarticle}
\usepackage{hyperref}
\hypersetup{colorlinks=true,bookmarksnumbered=true, bookmarksopen=true}
\texorpdfstring{\corref{cor}}{}
\usepackage{newtxtext, newtxmath}
\usepackage{amsmath,amsfonts}
\usepackage{bm}
\usepackage{booktabs}
\usepackage{algorithm}
\usepackage[noend]{algpseudocode}
\usepackage{paralist}
\usepackage{flushend}

\graphicspath{{figs/}}

\newcommand{\refsec}[1]{Section~\ref{#1}}
\newcommand{\reffig}[1]{Fig.~\ref{#1}}
\newcommand{\reftab}[1]{Table~\ref{#1}}

\newcommand{\refeqn}[1]{Equation~(\ref{#1})}

\journal{Astronomy and Computing}

\begin{document}

\begin{frontmatter}

\title{{Robust Gaussian Process Regression Based on Iterative Trimming}}

\author[addr1,mail1]{Zhao-Zhou Li }\corref{corr1}
\cortext[corr1]{Corresponding author}
\author[addr2,addr3,mail2]{Lu Li}
\author[addr2,addr4]{Zhengyi Shao}
\address[addr1]{Department of Astronomy, School of Physics and
  Astronomy, Shanghai Jiao Tong University, 955 Jianchuan Road,
  Shanghai 200240, China}
\address[addr2]{Key Laboratory for Research in Galaxies and Cosmology, 
  Shanghai Astronomical Observatory, Chinese Academy of Sciences,
  80 Nandan Road, Shanghai 200030, China}
\address[addr3]{University of the Chinese Academy of Sciences, No.19A Yuquan Road, Beijing 100049, China}
\address[addr4]{Key Lab for Astrophysics, Shanghai 200234, China}
\fntext[mail1]{\href{mailto:lizz.astro@gmail.com}{lizz.astro@gmail.com}}
\fntext[mail2]{\href{mailto:lilu@shao.ac.cn}{lilu@shao.ac.cn}}

\begin{abstract}

The Gaussian process (GP) regression
can be severely biased when the data are contaminated by outliers.
This paper presents a new robust GP regression algorithm that
iteratively trims the most extreme data points.
While the new algorithm retains the attractive properties of the standard GP
as a nonparametric and flexible regression method,
it can greatly improve the model accuracy for contaminated data even in the presence of extreme or abundant outliers.
It is also easier to implement compared with previous robust GP variants that rely on approximate inference.
Applied to a wide range of experiments with different contamination levels,
the proposed method significantly outperforms the standard GP
and the popular robust GP variant with the Student-$t$ likelihood in most test cases.
In addition, as a practical example in the astrophysical study, 
we show that this method can precisely determine the main-sequence ridge line in the color-magnitude diagram of star clusters.

\end{abstract}

\begin{keyword}
    Gaussian process \sep
    robust regression \sep
    outlier detection \sep
    ridge line \sep
    star clusters
\end{keyword}

\end{frontmatter}

\section*{Highlights}
\begin{compactitem}
  \item Nonparametric, flexible, and robust regression based on Gaussian process.
  \item Outperforming the robust Gaussian process with Student-$t$ likelihood significantly in many test cases.
  \item Easy to implement and computationally tractable.
  \item Practical example in the astrophysical study.
\end{compactitem}
\
\hrule

\section{Introduction}
\label{sec:intro}

There has been increasing interest in the Gaussian process (GP) regression \citep{Rasmussen2005a} 
method in both scientific research and industry applications.
As a non-parametric method, GP is completely data-driven. It does
not assume any explicit functional form between variables,
which is particularly attractive in the big data era.
Moreover, GP provides a Bayesian framework which can naturally
characterize prior and posterior distributions over functions.
As the basis of Bayesian optimization,
GP can serve as a probabilistic surrogate model for problems that demand sample efficiency.
For example, the cosmological emulators based on GP (e.g., \citealt{Heitmann2009a,McClintock2019a})
can make precise predictions in large parameter space 
using merely a finite set of numerical simulations.
This technique can save a substantial investment of time and resources
because each simulation is highly computation demanding.

\begin{figure}[t]
  \centering
  \includegraphics[width=0.48\textwidth]{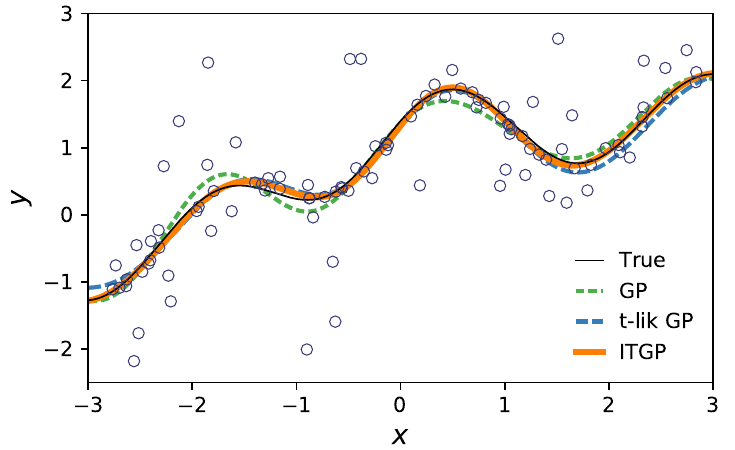}
  \vspace{-1.75em}
  \caption{%
   Illustration of the performance of three GP regression methods in the presence of outliers.
   45\% of the sample is corrupted with noises ten times that of the main sample
   (see Case ``abundant'' in \refsec{sec:neal_data} for details).
   The model prediction from the standard GP (green) is significantly biased,
   making robust GP methods necessary for better accuracy.
   Compared with the Student-$t$ likelihood GP (blue),
   the proposed ITGP (orange) shows better performance in this test case.
  }
  \label{fig:example_neal}
\end{figure}

GP can naturally handle the noises that are assumed to follow normal distributions.
However, predictions under this assumption are highly susceptible 
to the presence of extreme observations in data, the so-called outliers.
Outliers are usually generated by mechanisms different from the main sample, for example,
failure in measurement or calculation (e.g., broken sensor),
external factors (e.g., cosmic ray in astronomical images), 
and insufficient explanatory power of the model (e.g., binary stars for single stellar population models).
Their presence can make the estimate deviate substantially from the expected value,
as illustrated in \reffig{fig:example_neal}.
Therefore, robust regression techniques become necessary in such cases
(see \citealt{Huber2009} and \citealt{Maronna2019} for general reviews on robust statistics).

Several robust variants of GP regression have been proposed.
The most popular approach is to use alternative observation models that are less sensitive to extreme values,
for example, the heavy-tailed distributions such as
Student-$t$ \citep{Neal1997,Vanhatalo2009,Jylanki2011,Ranjan2016} or Laplace \citep{Kuss2006} distribution,
mixture of multiple Gaussians \citep{Kuss2006,Stegle2008,Ross2013a},
and input dependent noise model \citep{Goldberg1998,Naish-Guzman2007,Almosallam2017}.
However, unlike the Gaussian noise case, GPs with these models become analytically intractable,
hence relying on approximate inference techniques, e.g., 
variational approximation and expectation propagation,
which are often challenging to understand and implement.
More importantly, they may still suffer from the influence of outliers to some extent, 
as shown in \reffig{fig:example_neal} for GP with Student-$t$ likelihood ($t$-lik).

Another sensible strategy is to trim the outliers with large deviations from the GP model prediction,
then rerun GP on the purified sample.
However, noting that the initial model prediction itself is already affected by outliers,
the consequent residuals can not faithfully reveal the true deviations,
possibly leading to severe bias.
While this effect can be mitigated using iterations (e.g., \citealt{Wang2017c}),
it is a nontrivial task to figure out a general procedure that achieves
both high robustness and high statistical efficiency with the minimal possible computation cost.

This work proposes a new robust regression method based on the standard GP and iterative trimming (denoted as ITGP),
which is easy to implement and computationally tractable.
As shown in this paper,
while the new method retains the GP's attractive properties
as a nonparametric and flexible regression method,
it shows robust performance in the presence of outliers
and outperforms the popular $t$-lik GP significantly in many cases.

The rest of the paper is organized as follows.
In \refsec{sec:robustgp},
we first provide some preliminaries of the GP regression,
then present the proposed ITGP algorithm and corresponding practical guidance.
We compare the performance of ITGP and other methods for a variety of test cases in \refsec{sec:experiment}
and summarize in \refsec{sec:conclusion}.

Alongside this paper, 
we have released our open-source implementation of ITGP online (\url{https://github.com/syrte/robustgp/}),
which is written in \textsc{Python} based on the public GP package \textsc{GPy}.

\section{Robust Gaussian Process Regression}
\label{sec:robustgp}

\subsection{Basis of Gaussian process}
\label{sec:GP}

We consider a regression problem 
\begin{equation}
y = f (\bm{x}) + \epsilon,
\end{equation}
where the observable (aka response or target) $y$ is the summation of 
an underlying model, $f(\bm{x}): \mathbb{R}^d \rightarrow \mathbb{R}$,
and an observation noise $\epsilon$.
The object is to infer the latent function $f (\bm{x})$
from a dataset $\mathcal{D}=\{\bm{x}_i, y_i\}_{i=1}^n$, where $n$ is the sample size.

The Gaussian process (GP) provides a flexible prior on a family of functions.
If the underlying model $f$ is a realization of a GP
with a mean function $m(\bm{x})$ and a kernel function $k(\bm{x}, \bm{x}')$,
then the function values at arbitrary subset of locations,
$\bm{f}= \{ f (\bm{x}_i) \}_{i = 1}^n$, follow a multivariate Gaussian distribution,
\begin{equation}
  p (\bm{f}) =\mathcal{N} (\bm{f} \mid \bm{\mu}, \bm{\Sigma}).
\end{equation}
Here the mean vector and covariance matrix are determined by
$\bm{\mu}_i = m (\bm{x}_i)$ and $\bm{\Sigma}_{i,j} = k (\bm{x}_i,\, \bm{x}_j)$ respectively.
Conventionally, people use a zero mean function
and an analytical kernel function that is controlled by a set of hyper-parameters $\Theta$
(see \citealt{Duvenaud2014} for discussion on choice of kernel).

The standard GP assumes that 
the observation noise $\epsilon$ also follows a Gaussian distribution,
then $\bm{y}=\{ y_i \}_{i = 1}^n$ is still Gaussian and can be reinterpreted by adding a white noise term in the kernel $k$,
see \refeqn{eq:se} for an example.
It is particularly convenient because the inference is analytically tractable.

Given a training sample, $\mathcal{D}=\{\bm{x}_i, y_i\}_{i=1}^n$,
one can first infer the optimal hyper-parameters $\Theta$ for the kernel
by maximizing the likelihood $p(\bm{y}|\{\bm{x}_i\}, \Theta)$,
then use the conditional distribution to derive the posterior prediction
$p(f_\ast|\mathcal{D}, \Theta)$ at any new point $\bm{x}_\ast$,
including the mean $\hat f_\ast=\mathbb{E}[f_\ast]$ and the variance $\sigma^2_\ast=\mathrm{var} [f_\ast]$.
See \citet{Rasmussen2005a} for details. 

As mentioned in the introduction,
inferences with GPs are susceptible to outliers, which return extremely low likelihood under the Gaussian distribution.
To alleviate their influence,
one can keep $f$ as a realization of the GP
but use alternative observation models for $\epsilon$, e.g., Student-$t$ distribution,
whose long tail can lower the significance of the points with large deviation.
However, note that $p(y)$,
as the convolution of a Gaussian and a Student-$t$ distribution,
does not have an explicit form.
Therefore, the likelihood and the conditional posterior
are no longer analytically tractable hence requiring approximate inference techniques,
which are generally more challenging in both implementation and computation.

In the following, we present our new method ITGP,
which aims to sustain the simplicity of the standard GP
and strengthen the robustness simultaneously.

\subsection{Robust Gaussian process with iterative trimming}
\label{sec:algorithm}

\begin{figure}[hbt]
  \centering
  \includegraphics[width=0.48\textwidth]{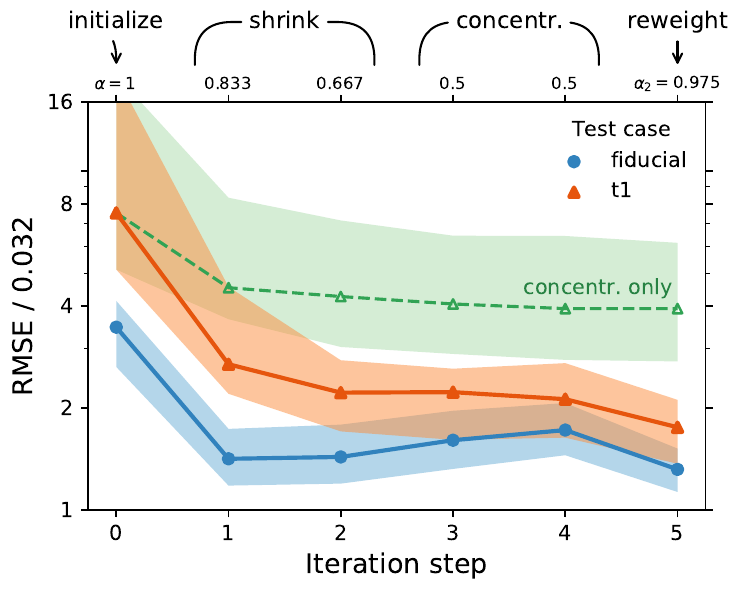}
  \vspace{-2em}
  \caption{%
  Illustration of the iterative trimming with two test cases.
  The solid curves show the median 
  of the root-mean-square error (RMSE, scaled by 0.032 for clarity) 
  of the prediction as a function of iteration steps,
  and the shaded bands show the interquartiles.
  Each case contains 50 datasets, and each dataset contains 100 data points 
  (see \refsec{sec:neal_data}).
  For each dataset and at each step, 
  a proportion (characterized by $\alpha$) of points with the largest residuals is removed,
  then a GP is retrained with the trimmed sample.
  The shrinking stage can prevent premature convergence and improve model accuracy,
  while the reweighting step can increase the statistical efficiency.
  For comparison, 
  the dashed curve shows results for Case ``$t_1$'' obtained by concentrating without shrinking and reweighting stages.
  }
  \label{fig:iter}
\end{figure}

The main idea of ITGP is to iteratively trim a proportion of
the points with the largest absolute residuals,
so that the remaining sample can better describe the bulk pattern of the data.
This method is summarized in Algorithm \ref{alg:robustgp3} and illustrated in \reffig{fig:iter}.
Training an ITGP model consists of three stages: shrinking, concentrating, and reweighting.

\begin{algorithm}[hbt]
\caption{Iterative trimming Gaussian process}
\label{alg:robustgp3}

\begin{algorithmic}[1]

\Function{ITGP}{$\bm{X}, \bm{y}, \alpha_1=0.5, \alpha_2=0.975, n_\mathrm{sh}=2, n_\mathrm{cc}=2$}
\Statex \hspace{2mm} $\triangleright$ \textit{Input}:
    training sample, $\bm{X}$ and $\bm{y}$, 
    trimming param, $\alpha_1$, 
    reweighting param, $\alpha_2$,
    number of shrinking and concentrating iterations, $n_\mathrm{sh}$ and $n_\mathrm{cc}$
\Statex \hspace{2mm} $\triangleright$ \textit{Output}:
    trimmed sample, $\bm{X}_I$ and $\bm{y}_I$, trained GP hyper params, $\Theta$,
    consistency factor, $c$

    \Statex
    \Statex \hspace{2mm} $\triangleright$ \textit{Shrinking} and \textit{concentrating} 
    \For{$j\gets 0,\, n_\mathrm{sh}+n_\mathrm{cc}$}
        \If {$j=0$}
            \State $I \gets \{ i \mid \bm{x}_i \in \bm{X} \}$
                \Comment start with full sample
            \State $c \gets 1$
        \Else
            \If {$j \leq n_\mathrm{sh}$}
                \Comment{{reduce} $\alpha$ from 1 to $\alpha_1$}
                \State $\alpha \gets 1 - (1-\alpha_1)\cdot j/(n_\mathrm{sh} + 1)$
            \Else
                \State $\alpha \gets \alpha_1$
            \EndIf 

            \State ${I} \gets \{ i \mid r_i \leq \alpha\text{-quantile}\,(\bm{r}) \}$
                \Comment{trim}

            \State $c \gets \alpha / F_{\chi^2_3}\big(\chi^2_{1,\alpha}\big)$ 
                \Comment{consistency factor}
        \EndIf

        \State $\Theta \gets \text{gp\_optimize}\,(\bm{X}_{I}, \bm{y}_{I})$ 
            \Comment train hyper params
        \Statex
            \Comment $\bm{X}_I := \{\bm{x}_i \in \bm{X}\}_{i\in I}$

        \State $\bm{\mu}, \bm{\sigma}^2 \gets \text{gp\_predict}\,(\bm{X} \mid \bm{X}_I, \bm{y}_I, \Theta)$
            \Comment{mean, var}
        \State $\bm{r} \gets |\bm{y} - \bm{\mu}|\,/\,(\bm{\sigma}\sqrt{c})$
            \Comment{normalized residual}
    \EndFor

    \Statex
    \Statex \hspace{2mm} $\triangleright$ \textit{Reweighting} 
        \State ${I} \gets \{ i \mid r_i^2 \leq {\chi^2_{1,\alpha_2}} \}$
            \Comment{retrim}
        \State $c \gets \alpha_2 / F_{\chi^2_3}\big(\chi^2_{1,\alpha_2}\big)$
            \Comment{consistency factor}

        \State $\Theta \gets \text{gp\_optimize}\,(\bm{X}_{I}, \bm{y}_{I})$
            \Comment retrain
        \Statex
        \State \Return  $\bm{X}_I, \bm{y}_I, \Theta, c$

\EndFunction
\end{algorithmic}
\end{algorithm}

\textbullet~\textit{Shrinking and concentrating} stages (lines 2--15). The procedure is as follows.
\begin{compactenum}
\item
First train the standard GP with the full sample $\{\bm{x}_i, y_i\}_{i=1}^{n}$,
predict the mean $\hat f_i$ and variance $\sigma^2_i=\mathrm{var}[\hat y_i]$ for each point,
and calculate corresponding normalized residual, $r'_i=|y_i-\hat f_i|/\sigma_i$.
\item
Retrain the GP using the $\lceil \alpha n \rceil$ points with 
the smallest residuals $r'$ and update the predictions $\{\hat f_i, \sigma_i, r'_i\}_{i=1}^{n}$.
\item
Repeat Step 2 
for $n_\mathrm{sh}+n_\mathrm{cc}$ times.
Let the preserving fraction $\alpha$
shrink from 1 to $\alpha_1$ gradually in the first $n_\mathrm{sh}$ iterations (shrinking stage)
and remain constant for the next $n_\mathrm{cc}$ iterations (concentrating stage).
Here $\alpha_1$, $n_\mathrm{sh}$, and $n_\mathrm{cc}$ are ITGP parameters.
\end{compactenum}

Note that the $\lceil \alpha n \rceil$ points are always selected from the full sample
so that a point inappropriately discarded in an earlier iteration could be brought back later.
For the first few iterations, the predicted mean might deviate from the underlying function 
substantially due to the influential outliers.
However, as shown in \reffig{fig:iter},
the deviation decreases fast because we always dispose of the outermost points in each iteration step.
We also find that a non-zero $n_\mathrm{sh}$ can 
effectively prevent premature convergence and significantly improve the model accuracy.

It is known that the variance of a $(1-\alpha)$-trimmed sample underestimates 
the actual variance of the underlying sample by a \textit{consistency factor}, $c$
(\citealt{Croux1999}, see also \citealt{Pison2002} for the finite-sample correction).
Under normality,
\begin{equation}
    c = \alpha / F_{\chi^2_{3}} \big(\chi^2_{1,\alpha}\big),
\end{equation}
where $F_{\chi^2_{3}}$ is the cumulative distribution function of the $\chi_{k=3}^2$ distribution (3-DoF $\chi^2$) 
and $\chi^2_{1,\alpha}$ is the $\alpha$-quantile of the $\chi_{k=1}^2$ distribution.
As reference, $c=2.65^2$, $1.65^2$ for $\alpha=0.5$, 0.75 respectively.
Therefore, the \textit{corrected} normalized residual writes
\begin{equation}
r_i = |y-\hat f_i|/(\sigma_i \sqrt{c}),
\end{equation}
where $\sigma^2_i$ is the variance predicted by the GP trained with the trimmed sample.

\textbullet~\textit{Reweighting} stage (lines 16--18).
To increase the statistical efficiency, one can further apply a one-step reweighting.
A simple yet effective choice \citep[e.g.,][]{Rousseeuw1987} is to remove the data points with $r_i^2> \chi^2_{1,\alpha_2}$
and retrain the GP with the remaining sample.
Under normality, this amounts to discarding a proportion of about $1 - \alpha_2$ of the points with
the largest absolute residuals.
It is customary to take $\alpha_2=0.95$ or 0.975 
(corresponding to $\chi^2_1=1.96^2$, $2.24^2$ and $c=1.15^2$, $1.08^2$).

Finally, once the ITGP is finalized,
one can then identify outliers by the \textit{corrected} normalized residuals $r_i$, 
e.g., label the data points with $r_i>3$ as 3$\sigma$ outliers (under normality).
Besides, the quantile-quantile (Q--Q) plot can also be helpful diagnostics.

\subsection{Practical guidance}
\label{sec:guidance}

To the best of our knowledge,
we recommend the following ITGP parameters for general problems:
$\alpha_1=0.5$, $n_\mathrm{sh}=2$, $n_\mathrm{cc}=2$, and $\alpha_2=0.975$.

We find the above values work remarkably well for all test cases presented in this work.
According to systematic experiments in the \ref{sec:hyperparam},
further tuning can bring slight improvements in particular cases,
but often at the price of higher computation cost or possibly worse performance for general problems.
Hence, our recommendation appears as a good compromise among robustness, efficiency, and computation cost.

It is worth emphasizing that one should not use any method blindly in practice.
Although ITGP shows general robustness against abundant and extreme outliers,
it is always good practice to clean the input data beforehand, use prior information when available,
and check if the model provides a reasonable summary of the bulk pattern in the data.
In particular, as a flexible nonlinear regression method,
ITGP could be biased by influential points near the boundary of the data coverage
when the local signal-to-noise ratio is low, see \refsec{sec:neal_data} and \reffig{fig:failure} for example.
Moreover, the first few iterations of ITGP inevitably suffer from masking and swamping effects, 
which might break down the whole iteration procedure in some rare cases.
Our experience suggests that 
taking one or several of the following measures can effectively prevent catastrophic failures
(e.g., dramatic oscillation in prediction) in difficult cases:
removing obvious outliers by truncating the data, 
setting reasonable bounds (even loose ones) for GP hyperparameters (i.e., the scale length and variance of kernels),
optimizing hyperparameters with a cold start (rather than values from the last iteration),
and using a larger $n_\mathrm{sh}$ (e.g., 5 or 10 rather than 2).

\subsection{Speed}
\label{sec:speed}

Given the $\mathcal{O}(n^3)$ complexity of GP,
naively, we expect the training time of ITGP with $n_\mathrm{sh}=2$ and $n_\mathrm{cc}=2$
to be about three times longer than the standard GP.
It is a lower limit, considering the additional computations for making prediction in each iteration.
According to the numerical experiments based on \textsc{GPy} in \refsec{sec:experiment},
the time cost of ITGP is typically 3.5 to 5 times that of the standard GP.

Obviously, the absolute time cost  depends on the implementation.
People have proposed many fast and scalable GP implementations (see \citealt{Liu2020a} for a recent review),
including exact solutions with fast and parallel linear algebra algorithms 
(e.g., \textsc{george}, \citealt{Ambikasaran2015};
\textsc{celerite}, \citealt{Foreman-Mackey2017})
and sparse approximations \citep[e.g., the Sparse Variational GP,][]{Hensman2013}.
Given that ITGP is completely based on the standard GP,
these techniques can be painlessly deployed to the ITGP for large datasets.

\subsection{Connection to LTS and other robust estimators}
\label{sec:lts}

There is a clear relevance between the proposed ITGP and the Least Trimmed Squares (LTS) estimators \citep{Rousseeuw1984a},
a popular robust (linear) regression method with a high breakdown value.
LTS looks for a subset of given size with the smallest mean squared residual from all possible combinations,
making itself a hard combinatorial problem.
\citet{Rousseeuw2006} proposed an efficient approximation algorithm (Fast-LTS),
which contains a similar concentrating procedure (the so-called C-step).
In a sense, ITGP can be seen as an approximate local solution of LTS.
The concept of the global LTS solution seems attractive;
however, it is nontrivial to properly define and efficiently solve it in the framework of GP regressions.
For example, the flexibility of GP even allows a solution to pass through every data point.
Therefore, the GP posterior probability might be a more appropriate target than the residuals.
We leave such exploration to future work.

The raw LTS estimator is long known for this low efficiency.
Besides the simple yet effective one-step reweighting adopted in this work,
there exist other robust estimators with high efficiency in the context of robust statistics
\citep[see][]{Yu2017,Maronna2019},
e.g., M-estimators \citep{Huber1964}, MM-estimators \citep{Yohai1987},
and REWLSE \citep{Gervini2002}.
It is probably worth incorporating such techniques into GP methods as well.
For example, \citet{Ramirez-Padron2021} recently proposed a robust weighted GP algorithm based on M-estimators,
which shows performance comparable to and sometimes better than $t$-lik GP.

It is also possible to combine different techniques, 
e.g., trimming based on preceding outlier detection via clustering algorithm (\citealt{Wang2017c})
or $t$-lik GP regression \citep{Martinez-Cantin2018a}.
Similarly, some advanced robust estimators like MM-estimators and REWLSE
require an initial robust estimate as input, where ITGP can be a good choice.

\section{Experiments with Synthetic Datasets} \label{sec:experiment}

In this section, we conduct a wide range of numerical experiments
to compare the performance of the proposed ITGP,
the standard GP, and the robust GP variant with Student-$t$ likelihood ($t$-lik).
The $t$-lik GP has been compared with many other robust GP variants and shows broadly similar performance
(e.g., \citealt{Kuss2006,Ranjan2016,Ramirez-Padron2021}),
hence serving as a good representative of existing methods.

The proposed ITGP is implemented in Python based on the public package \textsc{GPy},
and the remaining two methods are performed with \textsc{GPy} directly.
The hyperparameters are determined by the maximum a posterior (MAP) estimation
with the L-BFGS-B algorithm.%
\footnote{
The tests are performed on a computing node with 4 Intel Xeon Gold 6240 Processors (2.6GHz, $4\times18$ cores in total).
Array computations in \textsc{GPy} are automatically paralleled through \textsc{Numpy}.
}
We adopt the following configurations for the three methods respectively.
\begin{compactitem}
\item
  GP: the standard Gaussian process with Gaussian noise model.
  For reference, we also show the GP prediction (labeled as ``Ideal'')
  based on the purified sample with all contamination excluded.
  The latter presents the best possible performance in theory. 
\item
  $t$-lik GP: the Gaussian process with Student-$t$ noise model.
  We use the approximate posterior with the Laplace approximation,
  which shows performance very close to other implementations in terms of RMSE 
  \citep{Vanhatalo2009,Ranjan2016}.
  Changing the degree of freedom $\nu$ of a Student-$t$ distribution
  allows its shape to vary from Gaussian to heavy-tailed. 
\item
  ITGP: the iterative trimming Gaussian process with Gaussian noise model.
  We adopt the recommended parameters for all test cases: 
  $\alpha_1=0.5$, $n_\mathrm{sh}=2$, $n_\mathrm{cc}=2$, and $\alpha_2=0.975$.
\end{compactitem}

We present tests for two series of artificial datasets.
The first series of datasets was introduced in the seminal work of \citet{Neal1997};
the second was generated in a study on star clusters by \citet{Li2020a}.
For each test case, $50$ training datasets are generated and used to train the models separately.
The trained models are then evaluated on a noise-free test set with $m=2000$ data points.
Following \citet{Kuss2006}, to evaluate the performance, we report the root-mean-square error,
$\mathrm{RMSE}=(\frac{1}{m} \sum_{i=1}^{m} \Delta_i^2)^{1/2}$,
of the prediction residuals of each test set, $\{\Delta_i=f(x_{\ast,i})-\hat f(x_{\ast,i})\}_{i=1}^m$.
Consistent conclusions are obtained if using the mean absolute error instead.
We also provide the computation time for reference.
The average RMSE and time cost of each test case are summarized in 
\reftab{tab:neal_bench} and \ref{tab:cluster_bench}.

\subsection{Neal datasets} \label{sec:neal_data}

The \citet{Neal1997} datasets 
have been widely used as benchmarks for robust regression methods (e.g., \citealt{Vanhatalo2009,Ranjan2016}).
The underlying function $f(x)$ is given by 
\begin{equation}
  f (x) = 0.3 + 0.4 x + 0.5 \sin (2.7 x) + \frac{1.1}{1 + x^2}.
\end{equation}
To test the performance under different conditions,
we consider the following noise models for observation, $y=f(x)+\epsilon$.

\begin{compactitem}

\item Case 1: zero outliers,

  \qquad $\epsilon \sim \mathcal{N}(0,\ 0.1^2)$.

\item Case 2: rare outliers,

  \qquad $\epsilon \sim 0.95\ \mathcal{N}(0,\ 0.1^2) + \bm{0.05}\ \mathcal{N}(0,\ 1^2)$.

\item Case 3: fiducial case,

  \qquad $\epsilon \sim 0.85\ \mathcal{N}(0,\ 0.1^2) + 0.15\ \mathcal{N}(0,\ 1^2)$.

\item Case 4: abundant outliers,

  \qquad $\epsilon \sim 0.55\ \mathcal{N}(0,\ 0.1^2) + \bm{0.45}\ \mathcal{N}(0,\ 1^2)$.

\item Case 5: skewed outliers,

  \qquad $\epsilon \sim 0.85\ \mathcal{N}(0,\ 0.1^2) + 0.15\ \mathcal{N}(\bm{2},\ 1^2)$.

\item Case 6: extreme outliers,

  \qquad $\epsilon \sim 0.85\ \mathcal{N}(0,\ 0.1^2) + 0.15\ \mathcal{N}(0,\ \bm{5}^2)$.

\item Case 7: uniform outliers,

  \qquad $y \sim 0.7\ \mathcal{N}(f,\ 0.1^2) + \bm{0.3}\ \mathcal{U}[{-3},\ {3}]$.

\item Case 8: $t_3$ distribution with $\mathrm{var}[\epsilon]=3\times0.1^2$,

  \qquad $\epsilon \sim t\, (0,\ 0.1^2;\nu={3})$.

\item Case 9: $t_1$ (aka Cauchy) distribution with $\mathrm{var}[\epsilon]=\infty$,

  \qquad $\epsilon \sim t\, (0,\ 0.1^2;\nu={1})$.  

\end{compactitem}

\begin{figure}[bt]
  \centering
  \includegraphics[width=0.48\textwidth]{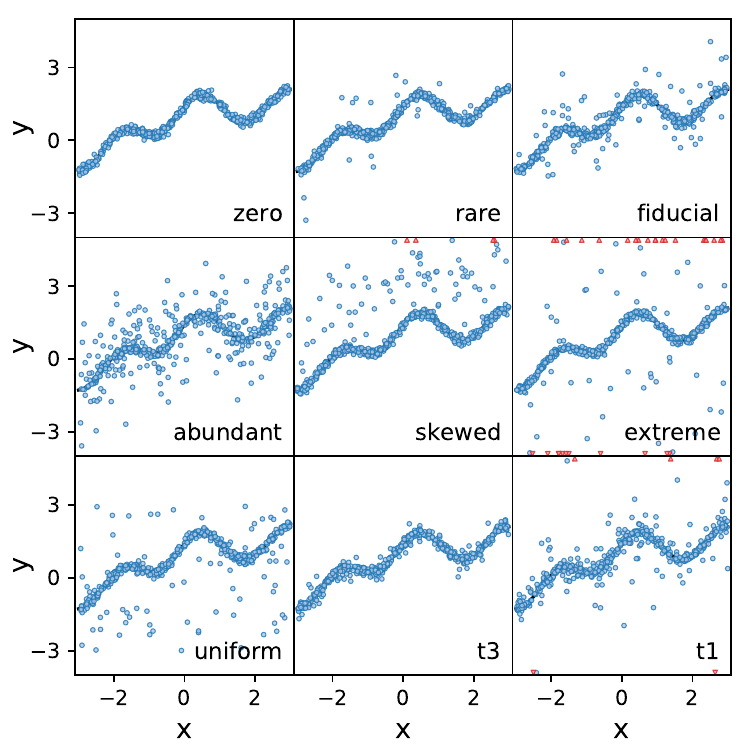}
  \vspace{-2em}
  \caption{%
  Example datasets for Cases 1--9.
  Some cases contain extreme outliers;
  those data points outside the box are indicated by triangles.
  }
  \label{fig:mock}
\end{figure}

Gaussian noise is assumed for the main samples in Cases 1--7,
and a Student-$t$ noise is assumed in Cases 8 and 9.
Moreover, Cases 2--6 are further contaminated by another Gaussian component with substantially larger 
($\times 10$ or more) scatter,
while Case 7 is contaminated by uniformly distributed outliers.
The difference between each case and the ``fiducial'' case is marked in bold.
Cases 4 and 9 are particularly challenging problems
for the excessive contamination fraction (45\%) and 
extremely distributed outliers ($\mathrm{var}[\epsilon]=\infty$).

Each test case contains 50 training samples,
and each sample contains $n=100$ data points randomly drawn from $x\in[-3, 3]$%
\footnote{
  It is slightly different from the original \citet{Neal1997} dataset, where $x$ was sampled from a normal distribution.
  It is because we are not interested in the possible leverage points in $x$ in this work.
}
(see \reffig{fig:mock} for examples).
We also generate larger samples with $n=500$ for comparison.
A noise-free test sample of $m=2000$ points is uniformly drawn in the same range.

We adopt the following kernel function for all tested methods,
\begin{equation}
k (x_i, x_j) = \sigma_\mathrm{k}^2 \exp \left( - d_{i j}^2/2 \right) + \sigma_\mathrm{w}^2 \delta_{i j},
\label{eq:se}
\end{equation}
where $d_{i j} = |x_i-x_j|/l_\mathrm{k}$.
The first term is a squared-exponential kernel,
which reflects the correlation between different locations.
The second term is a white noise kernel,
which captures the random errors due to observation noise.
In practice, a non-zero white noise term is generally recommended (even for $t$-lik GP)
to ensure valid matrix inversions.
The $t$-lik GP involves two more parameters, the degree of freedom (DoF), $\nu$, and the scale, $\sigma_t$, 
of the Student-$t$ observation model.
For each method,
the parameters $\Theta=\{l_\mathrm{k},\,\sigma_\mathrm{k},\,\sigma_w,\,[\nu,\,\sigma_t]\}$ are determined 
by maximizing the posterior probability of the training set.
Specifically, we update $\Theta$ in each iteration step of ITGP.

\begin{figure*}[htb!]
  \centering
  \includegraphics[width=0.95\textwidth]{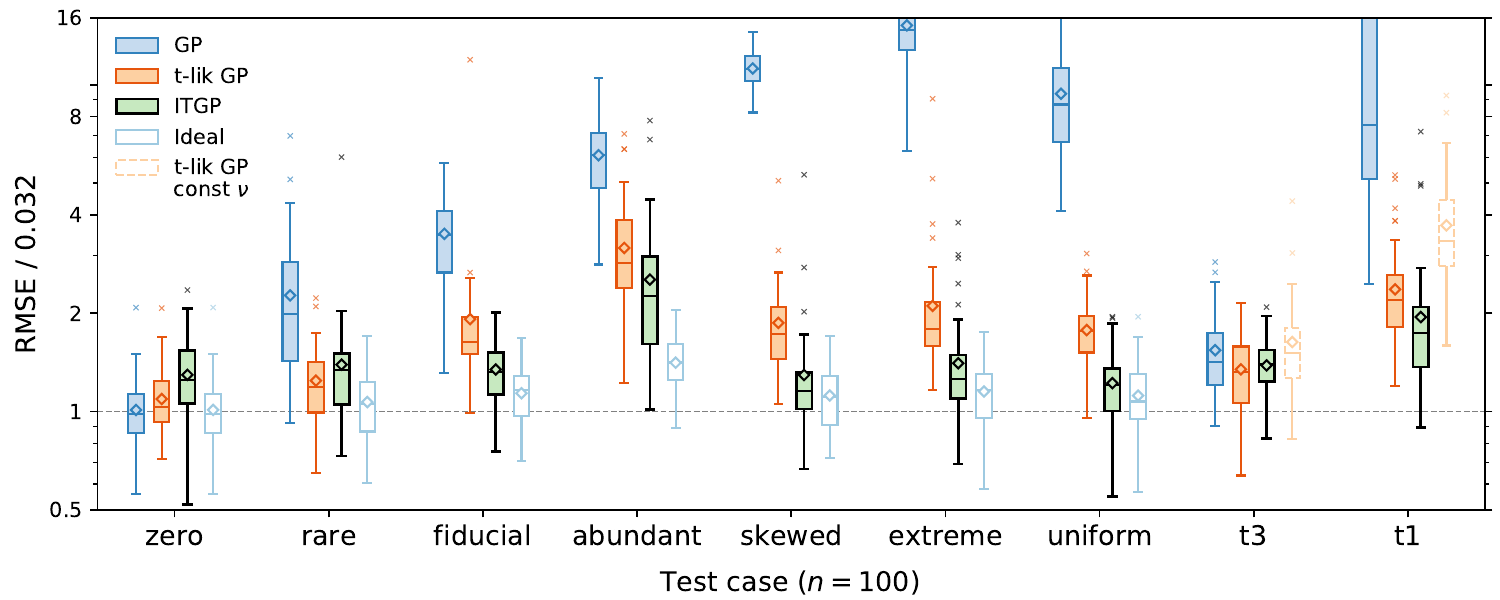}
  \includegraphics[width=0.95\textwidth]{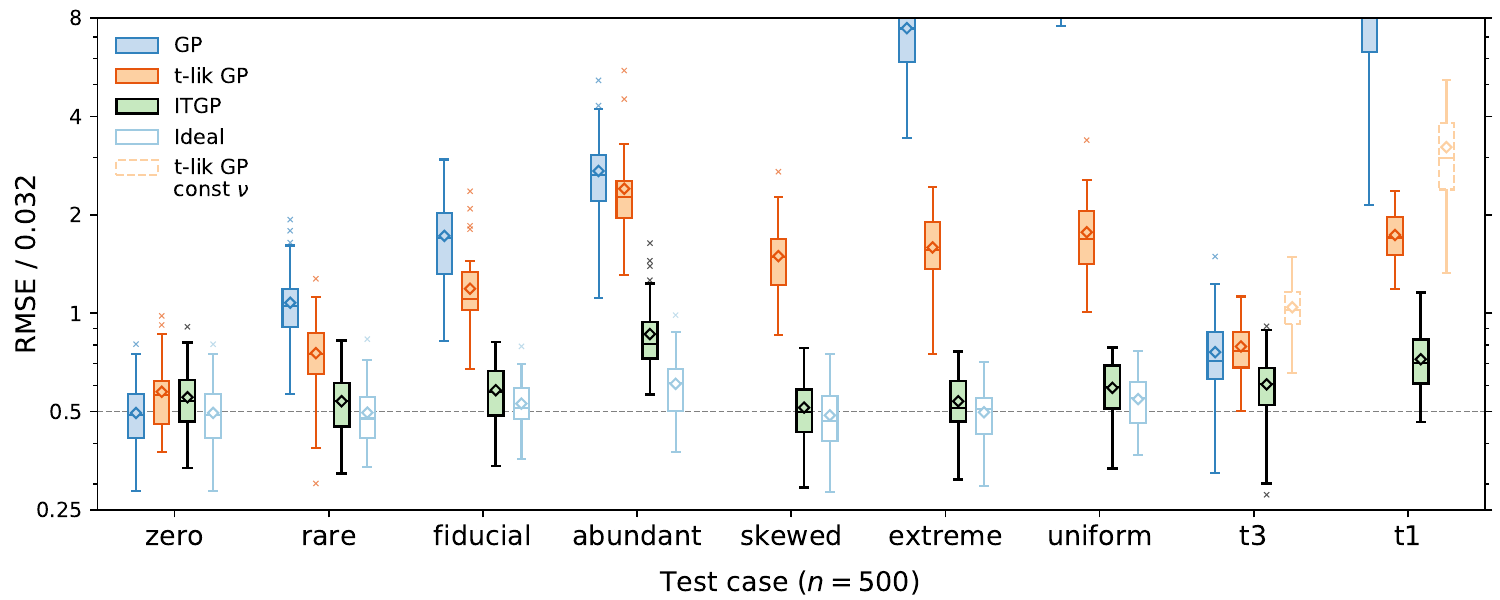}
  \vspace{-0.5em}
  \caption{%
  Performance comparison on Neal datasets.
  The root-mean-square errors (RMSE) are shown as box plots.
  In each box, the edges indicate the interquartile range,
  and the central line and the diamond symbol indicate the median and average, respectively. 
  }
  \label{fig:boxplot_neal}
\end{figure*}

\begin{figure*}[htb!]
  \centering
  \includegraphics[width=0.95\textwidth]{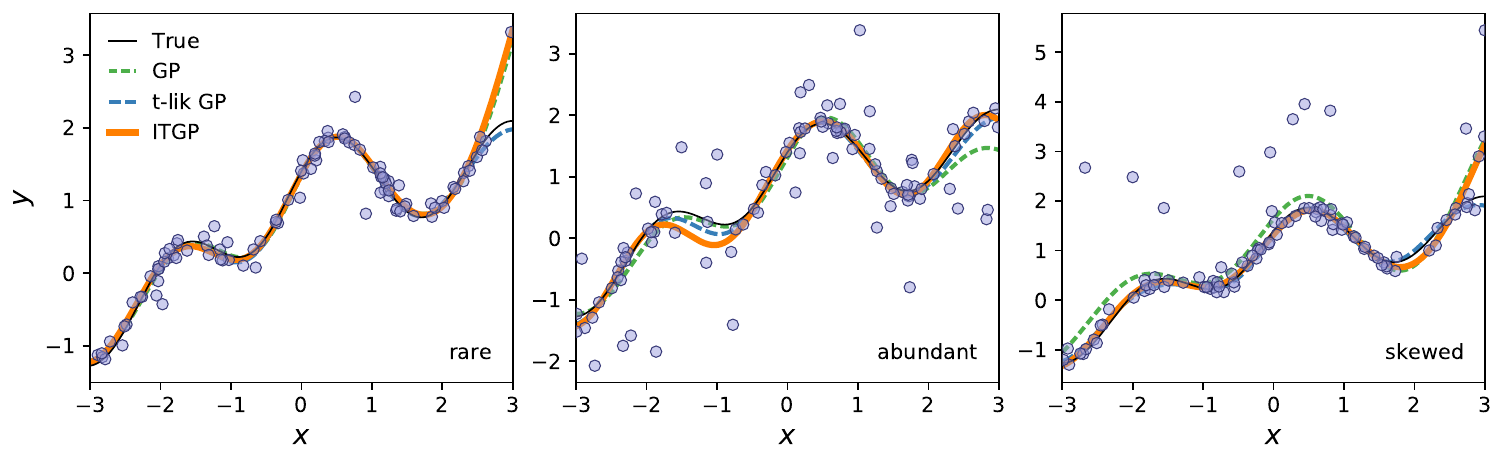}
  \vspace{-0.5em}
  \caption{%
  Three example datasets (with $n=100$) on which ITGP shows inferior performance.
  GP-based models can be biased in regions (mostly near the boundaries) where the local signal-to-noise ratio is low.
  This effect is less severe for larger data samples (e.g., $n=500$).
  }
  \label{fig:failure}
\end{figure*}

\reffig{fig:example_neal} shows model predictions for one example training set,
and \reffig{fig:boxplot_neal} presents box plots of RMSE
(see also \reftab{tab:neal_bench}).
For Cases 8 and 9, we also show the $t$-lik GP results with DoF fixed to the true value (3 and 1) for reference
(labeled as ``$t$-lik GP ($\nu$)'').

The standard GP works the best 
when the main sample is Gaussian and free of contamination (Case ``zero'')
and still provides good results with slight non-normality (Case ``$t_3$'').
However, it is highly susceptible to outliers in the other cases,
where $t$-lik GP and ITGP show significantly better performance.
$t$-lik GP seems comparable to or slightly better
than ITGP for low outlier fraction cases (Case ``zero'', ``rare'', and ``$t_3$'') 
when the sample size is small ($n=100$).
In other cases, ITGP significantly outperforms $t$-lik GP,
and more so for samples with $n=500$.

Remarkably, in several cases, e.g., ``skewed'', ``extreme'', and ``uniform'', 
the median RMSE of ITGP even approaches the ideal performance using purified samples (``Ideal GP'').
It is because outliers in these cases are more broadly dispersed and distinct from
the main sample (see \reffig{fig:mock}), hence in favor of trimming algorithms.
Moreover, ITGP's performance is better than or at least comparable to $t$-lik GP's 
even in the Student-$t$ noise cases (``$t_3$'' and ``$t_1$'').
It is perhaps because the long tail of Student-$t$ distribution leads to lower efficiency,
especially when $\nu$ is small.
We also find that when fixing $\nu$, even to the true value of the underlying sample, 
$t$-lik GP usually makes worse predictions due to less freedom in model fitting.

The RMSE of ITGP approximately reduces as $1/\sqrt{n}$,
while it does not apply to GP or $t$-lik GP in most cases.
For example, in Case ``skewed'', the mean RMSE of GP and $t$-lik GP
only reduces by 10\% and 20\%, respectively, when the sample size increases from 100 to 500.
The stars clusters in \refsec{sec:cluster_data} make an even more extreme example.
It is because the estimate's variance shrinks with larger $n$, but the bias does not.
This fact implies that ITGP has a much smaller bias than GP or $t$-lik GP for contaminated samples.

In \reffig{fig:boxplot_neal},
several datasets show much larger RMSE than others,
hence lying above the whiskers (third quartile plus 1.5 times interquartile range).
As demonstrated by examples in \reffig{fig:failure},
such inferior performance is mainly caused by low local signal-to-noise ratio.
Unlike linear regressions where data points are expected to follow a global pattern,
nonlinear regressions with highly flexible models such as GP (and its variants)
are more dependent on local data quality.
When the local outlier fraction is high 
due to heavy contamination or Poisson fluctuation,
outliers may have a chance to dominate the local model prediction.
This effect is more severe near the boundaries and for small samples.
In fact, 
if we leave aside the boundary regions,
the average RMSE within $[-2, 2]$ 
of these datasets is much smaller and indistinguishable from the others.
We also find that using more shrinking steps, e.g., $n_\mathrm{sh}=5$ or 10 rather than 2, 
may help alleviate this effect in certain cases (see \reffig{fig:hyperparam}b).

Finally, we provide a comparison of computation time in \reftab{tab:neal_bench}.
The time cost of ITGP is typically 3.5 to 5 times that of the standard GP
(see \refsec{sec:speed} for more discussion). 
The time cost of $t$-lik GP is significantly longer than GP,
in consistency with earlier benchmarks \citep{Jylanki2011,Ranjan2016}.
We avoid over-interpreting the comparison
because the running time strongly depends on the implementation
(e.g., inference approximation, optimizer, programming language, and parallelization)%
\footnote{
However, note that other $t$-lik GP implementations, e.g., the variational approximation 
and the scale-mixture representation with Markov chain Monte Carlo,
are much slower than the Laplace approximation adopted in this test.
}.
Nevertheless, the relative time between the standard GP and ITGP is always meaningful.

\begin{table}[hbt!]
\caption{%
  Comparison of different models for Neal datasets.
}
\label{tab:neal_bench}
\vspace{-0.5em}

\small
\begin{center}
\begin{tabular*}{1\columnwidth}{l @{\extracolsep{\fill}} cccc}
\toprule
     Sample size  &\multicolumn{2}{c}{$n=100$} & \multicolumn{2}{c}{$n=500$}\\
\midrule
                  & $\dfrac{\mathrm{RMSE}}{0.032}$ & ${\mathrm{Time}}/{\mathrm{s}}$   
                  & $\dfrac{\mathrm{RMSE}}{0.032}$ & ${\mathrm{Time}}/{\mathrm{s}}$ \\ 
\midrule
\multicolumn{5}{l}{Case 1: zero} \\
\cmidrule(lr){1-5}
  \quad GP              &   \textbf{1.0} &  0.069 &  \textbf{0.50} &   0.37 \\
  \quad $t$-lik GP        &            1.1 &    4.9 &           0.57 &     18 \\
  \quad ITGP            &            1.3 &   0.35 &           0.55 &    2.0 \\
  \quad \textit{Ideal}  &            1.0 &  0.070 &           0.50 &   0.37 \\
\midrule
\multicolumn{5}{l}{Case 2: rare} \\
\cmidrule(lr){1-5}
  \quad GP              &            2.3 &  0.073 &            1.1 &   0.44 \\
  \quad $t$-lik GP        &   \textbf{1.2} &    4.6 &           0.75 &     19 \\
  \quad ITGP            &            1.4 &   0.36 &  \textbf{0.54} &    2.0 \\
  \quad \textit{Ideal}  &            1.1 &  0.066 &           0.50 &   0.39 \\
\midrule
\multicolumn{5}{l}{Case 3: fiducial} \\
\cmidrule(lr){1-5}
  \quad GP              &            3.5 &  0.079 &            1.7 &   0.48 \\
  \quad $t$-lik GP        &            1.9 &    4.5 &            1.2 &     20 \\
  \quad ITGP            &   \textbf{1.3} &   0.35 &  \textbf{0.58} &    2.1 \\
  \quad \textit{Ideal}  &            1.1 &  0.057 &           0.53 &   0.39 \\
\midrule
\multicolumn{5}{l}{Case 4: abundant} \\
\cmidrule(lr){1-5}
  \quad GP              &            6.1 &  0.085 &            2.7 &   0.51 \\
  \quad $t$-lik GP        &            3.2 &    2.2 &            2.4 &     12 \\
  \quad ITGP            &   \textbf{2.5} &   0.36 &  \textbf{0.86} &    2.0 \\
  \quad \textit{Ideal}  &            1.4 &  0.033 &           0.61 &   0.28 \\
\midrule
\multicolumn{5}{l}{Case 5: skewed} \\
\cmidrule(lr){1-5}
  \quad GP              &             11 &  0.088 &            9.9 &   0.56 \\
  \quad $t$-lik GP        &            1.9 &    4.9 &            1.5 &     20 \\
  \quad ITGP            &   \textbf{1.3} &   0.35 &  \textbf{0.51} &    2.1 \\
  \quad \textit{Ideal}  &            1.1 &  0.058 &           0.49 &   0.39 \\
\midrule
\multicolumn{5}{l}{Case 6: extreme} \\
\cmidrule(lr){1-5}
  \quad GP              &             15 &   0.11 &            7.4 &   0.62 \\
  \quad $t$-lik GP        &            2.1 &    3.5 &            1.6 &     18 \\
  \quad ITGP            &   \textbf{1.4} &   0.38 &  \textbf{0.54} &    2.2 \\
  \quad \textit{Ideal}  &            1.2 &  0.058 &           0.50 &   0.39 \\
\midrule
\multicolumn{5}{l}{Case 7: uniform} \\
\cmidrule(lr){1-5}
  \quad GP              &             15 &   0.10 &             12 &   0.56 \\
  \quad $t$-lik GP        &            2.6 &    3.0 &            1.8 &     19 \\
  \quad ITGP            &   \textbf{1.6} &   0.37 &  \textbf{0.59} &    2.1 \\
  \quad \textit{Ideal}  &            1.3 &  0.046 &           0.55 &   0.33 \\
\midrule
\multicolumn{5}{l}{Case 8: $t_3$} \\
\cmidrule(lr){1-5}
  \quad GP              &            1.5 &  0.071 &           0.76 &   0.42 \\
  \quad $t$-lik GP        &   \textbf{1.3} &    3.6 &           0.79 &     15 \\
  \quad ITGP            &            1.4 &   0.35 &  \textbf{0.61} &    1.9 \\
  \quad $t$-lik GP ($\nu=3$)&          1.6 &    4.3 &            1.0 &     20 \\
\midrule
\multicolumn{5}{l}{Case 9: $t_1$} \\
\cmidrule(lr){1-5}
  \quad GP              &            110 &   0.10 &             79 &   0.76 \\
  \quad $t$-lik GP        &            2.4 &    3.2 &            1.7 &     15 \\
  \quad ITGP            &   \textbf{1.9} &   0.37 &  \textbf{0.72} &    2.3 \\
  \quad $t$-lik GP ($\nu=1$)&          3.7 &    3.5 &            3.2 &     16 \\
\bottomrule
\end{tabular*}
\end{center}
\vspace{-0.5em}
The smallest RMSE for each dataset is marked in bold.
\vspace{-0.5em}
\end{table}

\subsection{Star clusters} \label{sec:cluster_data}

\begin{figure}[hbt]
  \centering
  \includegraphics[width=0.48\textwidth]{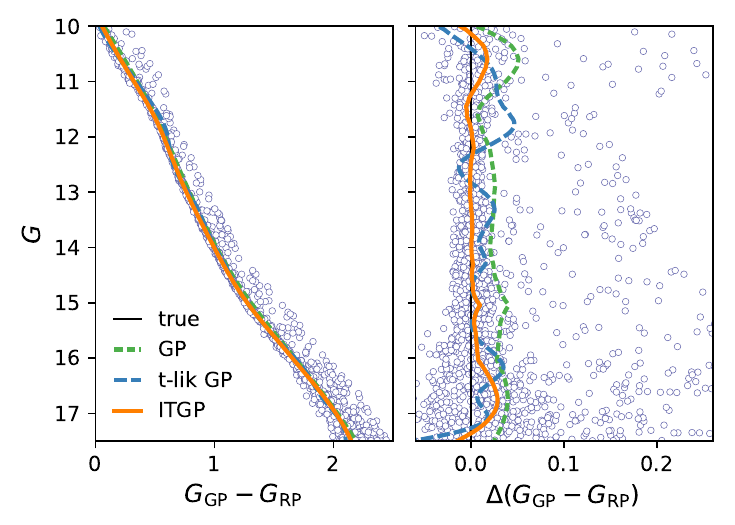}
  \vspace{-2em}
  \caption{%
  Color--magnitude diagram of a synthetic open cluster with 500 stars.
  Most single stars are distributed around the main-sequence ridge line,
  while the binary stars are extensively spread towards the upper right (more luminous and redder) side.
  The right panel shows the residual colors from the true ridge line.
  The curves present the model color, $G_\mathrm{GP}-G_\mathrm{RP}$, as a function of $G$ band magnitude
  predicted by three GP methods, respectively.
  }
  \label{fig:example_cluster}
\end{figure}

The stars in an open cluster are believed to 
share the same age, chemical composition, and many other physical properties.
Ideally, low-mass stars are expected to be located on a curve,
the so-called main sequence,
in the color--magnitude diagram (aka Hertzsprung--Russell diagram).
In observation, cluster stars actually follow an extended distribution peaked around the main sequence,
as shown in \reffig{fig:example_cluster}.
This extension is caused by observational errors and intrinsic scatters.
Moreover, the unresolved binary stars are spread broadly on the redder and brighter (right-top) side of the sequence,
serving as outliers in this problem.
A precise empirical determination of the main-sequence ridge line in observation is crucial to many studies, 
including inferring binary star properties and calibrating theoretical stellar models (see \citealt{Li2020a}).
As demonstrated by \citet{Li2020a},
the proposed ITGP can pinpoint the ridge line with high precision.

In the following, we present a benchmark test
using the realistic synthetic clusters generated by \citet{Li2020a}.
Assuming a stellar mass function and a binary mass-ratio distribution,
$2 \times 50$ open clusters are generated with the PARSEC stellar model \citep{Bressan2012},
each containing $500$ or $1000$ stars, $\sim\! 30\%$ of which are binary stars (outliers).
To mimic the observation from the Gaia mission,
we add a magnitude-dependent noise on each star,
making it a \textit{heteroscedastic} problem.
We aim to predict the color, $G_\mathrm{BP}-G_\mathrm{RP}$, as a function of the magnitude, $G$,
along the ridge line of the main sequence (see \reffig{fig:example_cluster}).

We first adopt the squared-exponential kernel (Equation~\ref{eq:se}) as the previous subsection.
\citet{Stein1999} argues that the weaker smoothness of the Matérn kernels 
makes them more realistic for modeling many physical processes.
Therefore, we also try the following kernels separately:
a composite Mat\'{e}rn 5/2 kernel,
\begin{equation}
k (x_i, x_j) = \sigma^2_{\mathrm{k}} \left( 1 + \sqrt{5} r_{i j} + 5r_{i j}/3 \right) \exp \left( - \sqrt{5} r_{i j} \right)
             + \sigma^2_\mathrm{w} \delta_{i j},
\end{equation}
and a composite Mat\'{e}rn 3/2 kernel,
\begin{equation}
k (x_i, x_j) = \sigma^2_{\mathrm{k}} \left( 1 + \sqrt{3} r_{i j} \right) \exp \left( - \sqrt{3} r_{i j} \right)
             + \sigma^2_\mathrm{w} \delta_{i j},
\end{equation}
where $r_{i j} = |x_i-x_j|/l_\mathrm{k}$.
Same as the Neal benchmarks, the parameters,
$\Theta=\{l_\mathrm{k},\,\sigma_\mathrm{k},\,\sigma_w,\,[\nu,\,\sigma_t]\}$
are determined by maximizing the posterior probability of the training sample.

\begin{figure}[bt]
  \centering
  \includegraphics[width=0.48\textwidth]{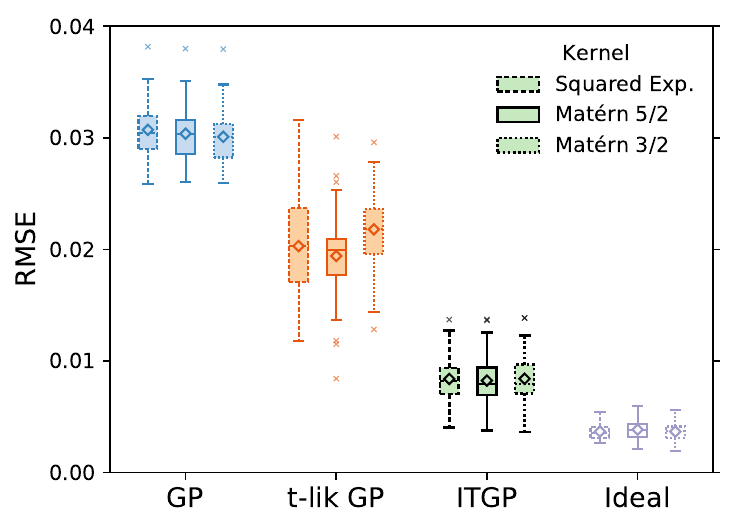}
  \vspace{-2em}
  \caption{%
  Performance comparison on synthetic star clusters. Each cluster contains $1000$ stars.
  }
  \label{fig:boxplot_cluster}
\end{figure}

\begin{table}[bt]
\caption{%
  Comparison of different models for star cluster datasets. 
}
\label{tab:cluster_bench}
\vspace{-0.5em}

\small
\begin{center}
\begin{tabular*}{1\columnwidth}{l @{\extracolsep{\fill}} cccc}
\toprule
     Sample size  &\multicolumn{2}{c}{$n=500$} & \multicolumn{2}{c}{$n=1000$}\\
\midrule
                  & ${\mathrm{RMSE}}$ & ${\mathrm{Time}}/{\mathrm{s}}$   
                  & ${\mathrm{RMSE}}$ & ${\mathrm{Time}}/{\mathrm{s}}$ \\ 
\midrule
\multicolumn{5}{l}{GP} \\
  \quad Squared Exp     &            0.031 &   0.51 &            0.031 &    2.2 \\
  \quad Matérn 5/2      &            0.031 &   0.61 &            0.030 &    2.2 \\
  \quad Matérn 3/2      &            0.031 &   0.81 &            0.030 &    2.9 \\
\multicolumn{5}{l}{$t$-lik GP} \\
  \quad Squared Exp     &            0.020 &     15 &            0.020 &     55 \\
  \quad Matérn 5/2      &            0.020 &     17 &            0.019 &     58 \\
  \quad Matérn 3/2      &            0.021 &     21 &            0.022 &     54 \\
\multicolumn{5}{l}{ITGP} \\
  \quad Squared Exp     &           0.0094 &    2.3 &           0.0084 &    6.4 \\
  \quad Matérn 5/2      &  \textbf{0.0092} &    2.8 &  \textbf{0.0082} &    7.4 \\
  \quad Matérn 3/2      &           0.0092 &    3.8 &           0.0084 &     10 \\
\multicolumn{5}{l}{\textit{Ideal}} \\
  \quad Squared Exp     &           0.0064 &   0.45 &           0.0037 &   0.89 \\
  \quad Matérn 5/2      &           0.0060 &   0.48 &           0.0038 &    1.1 \\
  \quad Matérn 3/2      &           0.0052 &   0.68 &           0.0037 &    1.5 \\
\bottomrule
\end{tabular*}
\end{center}
\vspace{-1em}
\end{table}

\reffig{fig:example_cluster} shows model predictions using the Mat\'{e}rn 5/2 kernel 
for one of the synthetic clusters,
while \reffig{fig:boxplot_cluster} presents the box plot for the overall performance
with different kernels
(see also \reftab{tab:cluster_bench}).
In general, the results are not sensitive to the choice of kernel,
though the Matérn 5/2 kernel seems slightly favored by t-lik GP.
The proposed ITGP again shows significantly better performance than GP and $t$-lik GP
in this problem, where the data is contaminated by abundant and skewed outliers (binary stars).

It is noteworthy that ITGP with a simple homoscedastic kernel provides remarkable performance 
even in this heteroscedastic problem.
Nevertheless, using a universal variance will trim more data points than optimal 
in the region where the noise is greater than average (e.g., at $G \simeq 17$),
which may lower the local efficiency and lead to bias.
It is probably why doubling the sample size only brings minor improvement in the RMSE of ITGP.
The results can be further improved by using heteroscedastic kernels in principle.
We leave such exploration to future work.

\section{Conclusion} \label{sec:conclusion}

This work has proposed a new robust regression algorithm based on the Gaussian process and iterative trimming (ITGP).
It greatly improves the model accuracy of GP in the presence of outliers
by iteratively removing the most extreme data points.
A novel shrinking procedure that gradually increases the trimming fraction
is introduced to prevent premature convergence,
and a one-step reweighting is used to increase the statistical efficiency.
The advantage of ITGP lies in its robustness, efficiency, ease of implementation, and
computational tractability.

Applied to a wide range of synthetic datasets with different contamination levels,
ITGP significantly outperforms the standard GP
and the popular robust GP variant with the Student-$t$ likelihood in most test cases,
including some particularly challenging problems
with excessive contamination fraction (45\%), extremely distributed outliers,
or moderate heteroskedasticity.

Though the optimal method always depends on the specific problem in principle, 
ITGP, nevertheless, shows remarkable performance as a general method,
thus ensuring wide application.
In addition, ITGP may serve as a good initial estimate for other advanced robust estimators
(e.g., \citealt{Yohai1987,Gervini2002})
for possible future improvement.

We have made our implementation of ITGP available online (\url{https://github.com/syrte/robustgp/}),
which is written in \textsc{Python} based on the public GP package \textsc{GPy}.
It is also straightforward to implement ITGP with other GP packages
for better speed and scalability.

\section*{Acknowledgments}

We thank Jiaxin Han and Hai Yu for helpful discussions
and the anonymous referees for constructive criticisms and suggestions.
This work is supported by National Key Basic R\&D Program of China (2018YFA0404504, 2019YFA0405501),
NSFC (11621303, 11873038, 11890691, 11973032, 12022307, U2031139), and 111 project (B20019). 
ZZL gratefully acknowledges the support of the MOE Key Lab for Particle Physics,
Astrophysics and Cosmology, Ministry of Education. 
The computation of this work is partly done on the \textsc{Gravity} supercomputer at the Department of Astronomy, Shanghai Jiao Tong University.

During the preparation of this work, we made use of the following open source software:
GPML\footnote{\url{http://www.gaussianprocess.org/gpml/code/}}, 
GPy\footnote{\url{https://github.com/SheffieldML/GPy}}, 
Jupyter\footnote{\url{https://jupyter.org/}}, 
Matplotlib\footnote{\url{https://matplotlib.org/}}, 
Numpy\footnote{\url{https://numpy.org/}}, 
Oct2Py\footnote{\url{https://github.com/blink1073/oct2py}}, 
Octave\footnote{\url{https://www.gnu.org/software/octave/}}, 
Python\footnote{\url{https://www.python.org/}}, 
and 
Scipy\footnote{\url{https://www.scipy.org/}}.

\appendix
\section{Choice of ITGP hyperparameters}
\label{sec:hyperparam}

\begin{figure*}[htb!]
  \centering
  \includegraphics[width=0.95\textwidth]{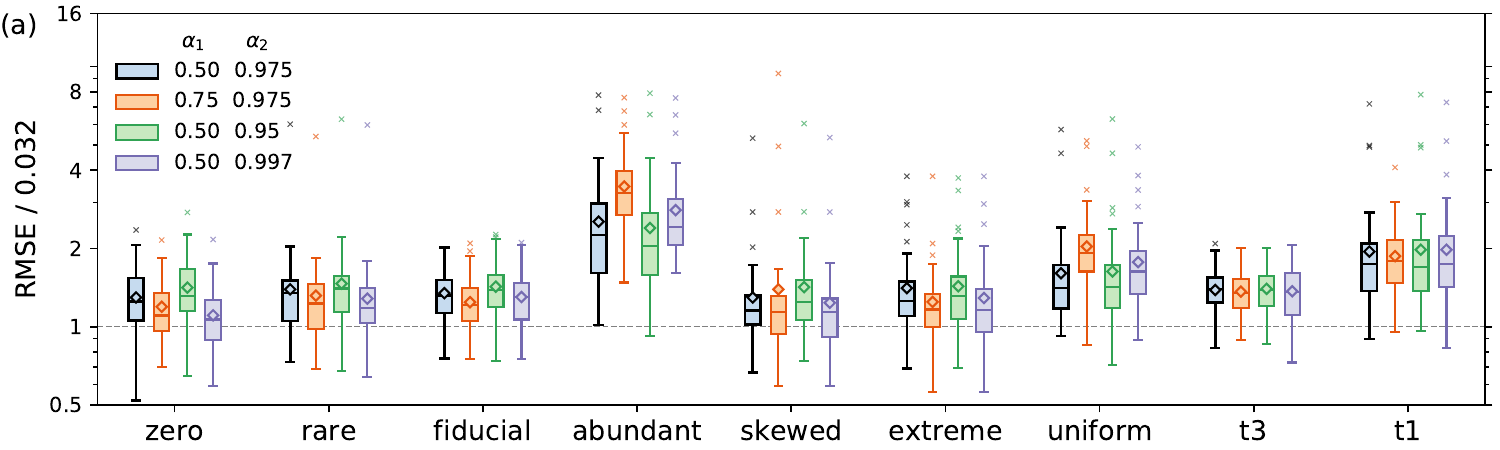}
  \includegraphics[width=0.95\textwidth]{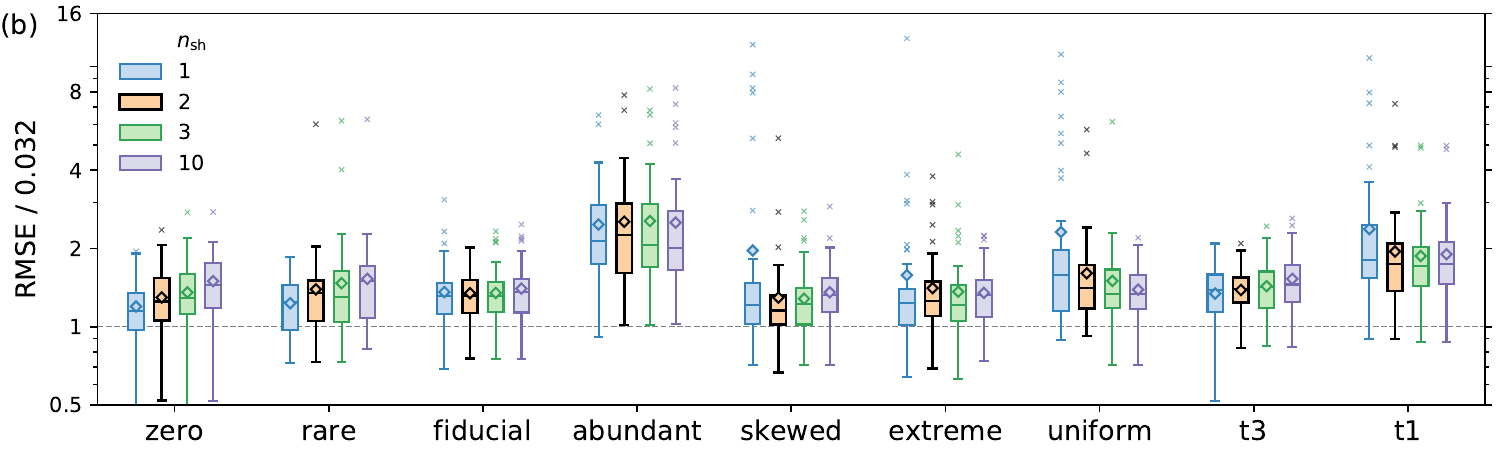}
  \includegraphics[width=0.95\textwidth]{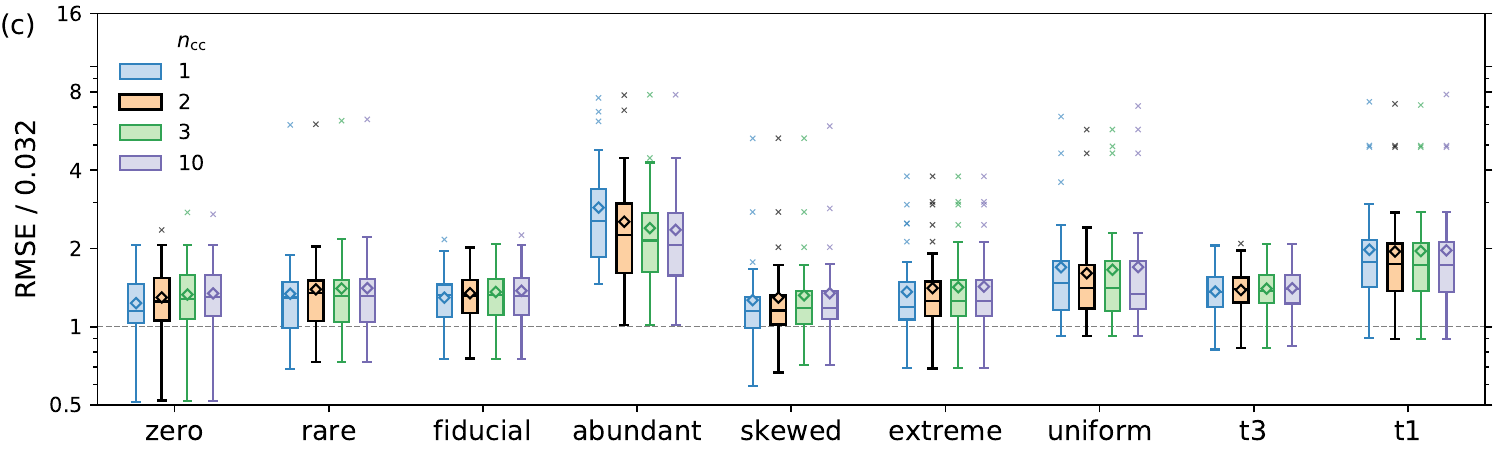}
  \includegraphics[width=0.95\textwidth]{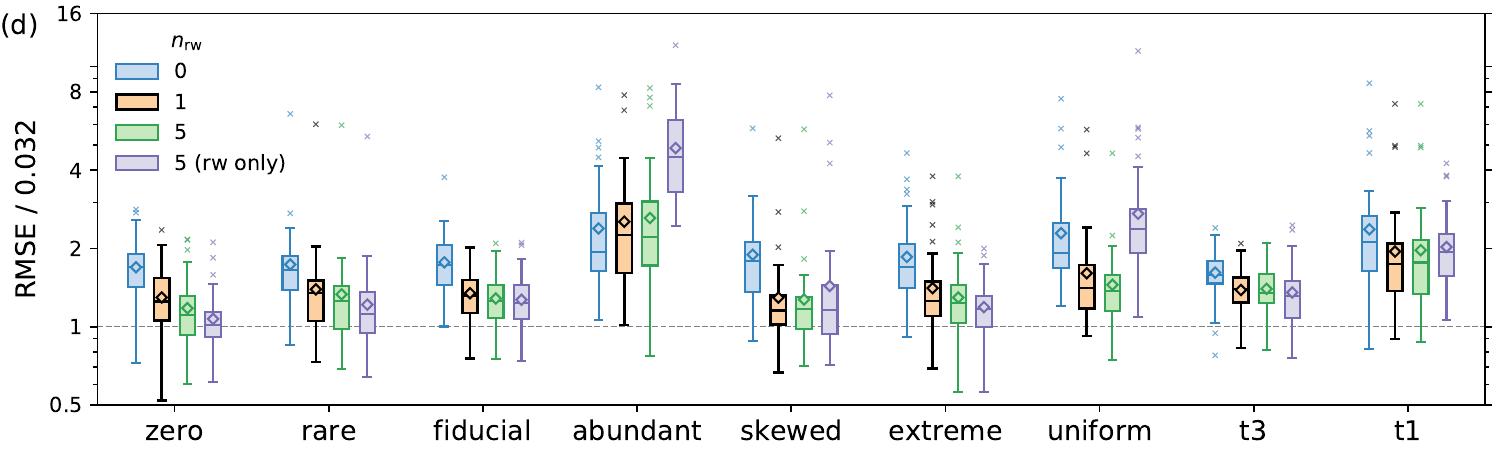}
  \caption{%
  Similar to \reffig{fig:boxplot_neal}, but showing the dependence of ITGP's performance on hyperparameters 
  for test cases with $n=100$.
  In each panel, one or several parameters have been changed from the fiducial values.
  The fiducial configuration
  ($\alpha_1=0.5$,  $\alpha_2=0.975$, $n_\mathrm{sh}=3$, $n_\mathrm{cc}=1$, and $n_\mathrm{rw}=1$)
  is marked by a black box in each test case.
  }
  \label{fig:hyperparam}
  \vspace{1em}
\end{figure*}

This appendix presents a series of numerical experiments
aiming to understand the ITGP hyperparameters and find the possible optimal choice.
We consider the following parameters:
    the trimming and reweighting parameters, $\alpha_1$ and $\alpha_2$,
    iteration numbers for shrinking, concentrating, and reweighting stages,
    $n_\mathrm{sh}$, $n_\mathrm{cc}$, and $n_\mathrm{rw}$.
We systematically change one or several of the above parameters and perform the 
same tests on the synthetic Neal datasets introduced in \refsec{sec:neal_data}.
The results with sample size $n=100$ are shown in \reffig{fig:hyperparam}.
We do not find any significant dependence of the following conclusions on the sample size.

In general, we find the fiducial values used in the main text,
$\alpha_1=0.5$,  $\alpha_2=0.975$, $n_\mathrm{sh}=3$, $n_\mathrm{cc}=2$,
and $n_\mathrm{rw}=1$ (one-step reweighting),
work remarkably well for all test cases.
Although the best configuration can depend on the specific problem,
the improvement from further tuning is usually limited.
Below we summarize some general expectations about hyperparameters and performance.

\textbullet~$\alpha_1$.
Using a larger $\alpha_1$, e.g., 0.75, can mildly increase the efficiency
as long as $\alpha_1 < 1 - \pi_\mathrm{outlier}$, where $\pi_\mathrm{outlier}$ is the outlier fraction (\reffig{fig:hyperparam}a).
However, if $\alpha_1$ is too large, 
e.g., $0.75$ for Case ``abundant'' and ``uniform'' ($\pi_\mathrm{outlier}=0.45$ and 0.3),
the performance may get significantly worsen due to the incomplete exclusion of influential outliers.
Therefore, 0.5 is recommended for maximum robustness,
unless one knows the upper limit of the outlier fraction in prior.

\textbullet~\textit{$n_\mathrm{sh}$}.
The shrinking steps can effectively prevent premature convergence (\reffig{fig:iter}).
While the average performance is not very sensitive to $n_\mathrm{sh}$ once $n_\mathrm{sh}\geq 2$ (\reffig{fig:hyperparam}b),
a larger $n_\mathrm{sh}$ might help improve the inferior performance
on some particular datasets (e.g., for Case ``extreme'').

\textbullet~\textit{$n_\mathrm{cc}$}.
The number of concentrating steps $n_\mathrm{cc}$ has little effect on ITGP performance in most cases (\reffig{fig:hyperparam}c).
The concentrating stage can approach or even achieve convergence within two or three steps (see also \reffig{fig:iter}),
though several more steps are generally needed for exact convergence when the sample is large.
Comparing with $n_\mathrm{cc}=2$, we find that allowing complete convergence makes little difference,
especially when a successive reweighting is added.
Larger $n_\mathrm{cc}$ is slightly favored when $\alpha_1$ is close to $1-\pi_\mathrm{outlier}$
(e.g., in Case ``abundant'').
A rule of thumb is to ensure at least 2 or 3 iterations (including shrinking steps) 
with $\alpha < 1-\pi_\mathrm{outlier}$ before the final reweighting.

\textbullet~\textit{$\alpha_2$}.
Using $\alpha_2=0.95$, 0.975, or 0.997 typically gives very similar results,
though one of them can be slightly favored depending on the nature of outliers (\reffig{fig:hyperparam}a).
Using even larger values is not recommended.
For an obvious example, it turns back to the standard GP with $\alpha_2=1$.

\textbullet~\textit{$n_\mathrm{rw}$}.
The one-step reweighting ($n_\mathrm{rw}=1$) can significantly improve the model precision 
if $\alpha_1 \ll 1 - \pi_\mathrm{outlier}$ but slightly worsen the result otherwise (\reffig{fig:hyperparam}d).
It is because the efficiency after the concentrating stage is already close to optimal
in the latter case (e.g.,  Case ``abundant'').
One may wonder what if performing iterative reweighting steps in a way similar to concentrating steps.
We find that using $n_\mathrm{rw}>1$ can slightly reduce RMSE in several cases 
at the cost of more computation (\reffig{fig:hyperparam}d).
Interestingly, a naive approach (labeled as ``rw only'')
using iterative reweighting without preceding shrinking and concentrating steps
(adopted by e.g., \citealt{Wang2017c})
also shows comparable performance in several cases.
However, this variant is significantly less favorable for heavily contaminated cases (``abundant'' and ``uniform'')
compared with the proposed ITGP.

In summary, our recommended parameters seem a good compromise among robustness, efficiency, and computation cost
for general problems.

\bibliographystyle{elsarticle-harv} 
\bibliography{robustgp}

\begin{thebibliography}{35}
\expandafter\ifx\csname natexlab\endcsname\relax\def\natexlab#1{#1}\fi
\providecommand{\url}[1]{\texttt{#1}}
\providecommand{\href}[2]{#2}
\providecommand{\path}[1]{#1}
\providecommand{\DOIprefix}{doi:}
\providecommand{\ArXivprefix}{arXiv:}
\providecommand{\URLprefix}{URL: }
\providecommand{\Pubmedprefix}{pmid:}
\providecommand{\doi}[1]{\href{http://dx.doi.org/#1}{\path{#1}}}
\providecommand{\Pubmed}[1]{\href{pmid:#1}{\path{#1}}}
\providecommand{\bibinfo}[2]{#2}
\ifx\xfnm\relax \def\xfnm[#1]{\unskip,\space#1}\fi
\bibitem[{Almosallam(2017)}]{Almosallam2017}
\bibinfo{author}{Almosallam, I.}, \bibinfo{year}{2017}.
\newblock \bibinfo{title}{Heteroscedastic {{Gaussian}} Processes for Uncertain
  and Incomplete Data}.
\newblock \bibinfo{type}{Ph.{{D}}.}. University of Oxford.
\bibitem[{{Ambikasaran} et~al.(2015){Ambikasaran}, {Foreman-Mackey},
  {Greengard}, {Hogg} and {O'Neil}}]{Ambikasaran2015}
\bibinfo{author}{{Ambikasaran}, S.}, \bibinfo{author}{{Foreman-Mackey}, D.},
  \bibinfo{author}{{Greengard}, L.}, \bibinfo{author}{{Hogg}, D.W.},
  \bibinfo{author}{{O'Neil}, M.}, \bibinfo{year}{2015}.
\newblock \bibinfo{title}{{Fast Direct Methods for Gaussian Processes}}.
\newblock \bibinfo{journal}{IEEE Transactions on Pattern Analysis and Machine
  Intelligence} \bibinfo{volume}{38}, \bibinfo{pages}{252}.
\newblock \DOIprefix\doi{10.1109/TPAMI.2015.2448083},
  \href{http://arxiv.org/abs/1403.6015}{{\tt arXiv:1403.6015}}.
\bibitem[{{Bressan} et~al.(2012){Bressan}, {Marigo}, {Girardi}, {Salasnich},
  {Dal Cero}, {Rubele} and {Nanni}}]{Bressan2012}
\bibinfo{author}{{Bressan}, A.}, \bibinfo{author}{{Marigo}, P.},
  \bibinfo{author}{{Girardi}, L.}, \bibinfo{author}{{Salasnich}, B.},
  \bibinfo{author}{{Dal Cero}, C.}, \bibinfo{author}{{Rubele}, S.},
  \bibinfo{author}{{Nanni}, A.}, \bibinfo{year}{2012}.
\newblock \bibinfo{title}{{PARSEC: stellar tracks and isochrones with the
  PAdova and TRieste Stellar Evolution Code}}.
\newblock \bibinfo{journal}{\mnras} \bibinfo{volume}{427},
  \bibinfo{pages}{127--145}.
\newblock \DOIprefix\doi{10.1111/j.1365-2966.2012.21948.x},
  \href{http://arxiv.org/abs/1208.4498}{{\tt arXiv:1208.4498}}.
\bibitem[{Croux and Haesbroeck(1999)}]{Croux1999}
\bibinfo{author}{Croux, C.}, \bibinfo{author}{Haesbroeck, G.},
  \bibinfo{year}{1999}.
\newblock \bibinfo{title}{Influence {{Function}} and {{Efficiency}} of the
  {{Minimum Covariance Determinant Scatter Matrix Estimator}}}.
\newblock \bibinfo{journal}{Journal of Multivariate Analysis}
  \bibinfo{volume}{71}, \bibinfo{pages}{161--190}.
\newblock \DOIprefix\doi{10.1006/jmva.1999.1839}.
\bibitem[{Duvenaud(2014)}]{Duvenaud2014}
\bibinfo{author}{Duvenaud, D.}, \bibinfo{year}{2014}.
\newblock \bibinfo{title}{Automatic Model Construction with {{Gaussian}}
  Processes}.
\newblock \bibinfo{type}{Thesis}. University of Cambridge.
\newblock \DOIprefix\doi{10.17863/CAM.14087}.
\bibitem[{{Foreman-Mackey} et~al.(2017){Foreman-Mackey}, {Agol}, {Ambikasaran}
  and {Angus}}]{Foreman-Mackey2017}
\bibinfo{author}{{Foreman-Mackey}, D.}, \bibinfo{author}{{Agol}, E.},
  \bibinfo{author}{{Ambikasaran}, S.}, \bibinfo{author}{{Angus}, R.},
  \bibinfo{year}{2017}.
\newblock \bibinfo{title}{{Fast and Scalable Gaussian Process Modeling with
  Applications to Astronomical Time Series}}.
\newblock \bibinfo{journal}{\aj} \bibinfo{volume}{154}, \bibinfo{pages}{220}.
\newblock \DOIprefix\doi{10.3847/1538-3881/aa9332},
  \href{http://arxiv.org/abs/1703.09710}{{\tt arXiv:1703.09710}}.
\bibitem[{Gervini and Yohai(2002)}]{Gervini2002}
\bibinfo{author}{Gervini, D.}, \bibinfo{author}{Yohai, V.J.},
  \bibinfo{year}{2002}.
\newblock \bibinfo{title}{A class of robust and fully efficient regression
  estimators}.
\newblock \bibinfo{journal}{Annals of Statistics} \bibinfo{volume}{30},
  \bibinfo{pages}{583--616}.
\newblock \DOIprefix\doi{10.1214/aos/1021379866}.
\bibitem[{Goldberg et~al.(1998)Goldberg, Williams and Bishop}]{Goldberg1998}
\bibinfo{author}{Goldberg, P.W.}, \bibinfo{author}{Williams, C.K.I.},
  \bibinfo{author}{Bishop, C.M.}, \bibinfo{year}{1998}.
\newblock \bibinfo{title}{Regression with input-dependent noise: A {{Gaussian}}
  process treatment}, in: \bibinfo{booktitle}{Proceedings of the 1997
  Conference on {{Advances}} in Neural Information Processing Systems 10},
  \bibinfo{publisher}{{MIT Press}}, \bibinfo{address}{{Cambridge, MA, USA}}.
  pp. \bibinfo{pages}{493--499}.
\bibitem[{{Heitmann} et~al.(2009){Heitmann}, {Higdon}, {White}, {Habib},
  {Williams}, {Lawrence} and {Wagner}}]{Heitmann2009a}
\bibinfo{author}{{Heitmann}, K.}, \bibinfo{author}{{Higdon}, D.},
  \bibinfo{author}{{White}, M.}, \bibinfo{author}{{Habib}, S.},
  \bibinfo{author}{{Williams}, B.J.}, \bibinfo{author}{{Lawrence}, E.},
  \bibinfo{author}{{Wagner}, C.}, \bibinfo{year}{2009}.
\newblock \bibinfo{title}{{The Coyote Universe. II. Cosmological Models and
  Precision Emulation of the Nonlinear Matter Power Spectrum}}.
\newblock \bibinfo{journal}{\apj} \bibinfo{volume}{705},
  \bibinfo{pages}{156--174}.
\newblock \DOIprefix\doi{10.1088/0004-637X/705/1/156},
  \href{http://arxiv.org/abs/0902.0429}{{\tt arXiv:0902.0429}}.
\bibitem[{Hensman et~al.(2013)Hensman, Fusi and Lawrence}]{Hensman2013}
\bibinfo{author}{Hensman, J.}, \bibinfo{author}{Fusi, N.},
  \bibinfo{author}{Lawrence, N.D.}, \bibinfo{year}{2013}.
\newblock \bibinfo{title}{Gaussian processes for {{Big}} data}, in:
  \bibinfo{booktitle}{Proceedings of the {{Twenty}}-{{Ninth Conference}} on
  {{Uncertainty}} in {{Artificial Intelligence}}}, \bibinfo{publisher}{{AUAI
  Press}}, \bibinfo{address}{{Bellevue, WA}}. pp. \bibinfo{pages}{282--290}.
\bibitem[{Huber(1964)}]{Huber1964}
\bibinfo{author}{Huber, P.J.}, \bibinfo{year}{1964}.
\newblock \bibinfo{title}{Robust {{Estimation}} of a {{Location Parameter}}}.
\newblock \bibinfo{journal}{The Annals of Mathematical Statistics}
  \bibinfo{volume}{35}, \bibinfo{pages}{73--101}.
\newblock \DOIprefix\doi{10.1214/aoms/1177703732}.
\bibitem[{Huber and Ronchetti(2009)}]{Huber2009}
\bibinfo{author}{Huber, P.J.}, \bibinfo{author}{Ronchetti, E.M.},
  \bibinfo{year}{2009}.
\newblock \bibinfo{title}{Robust Statistics}.
\newblock \bibinfo{publisher}{{John Wiley \& Sons, Inc.}}
\newblock \DOIprefix\doi{10.1002/9780470434697}.
\bibitem[{Jyl{\"a}nki et~al.(2011)Jyl{\"a}nki, Vanhatalo and
  Vehtari}]{Jylanki2011}
\bibinfo{author}{Jyl{\"a}nki, P.}, \bibinfo{author}{Vanhatalo, J.},
  \bibinfo{author}{Vehtari, A.}, \bibinfo{year}{2011}.
\newblock \bibinfo{title}{Robust {{Gaussian Process Regression}} with a
  {{Student}}-t {{Likelihood}}}.
\newblock \bibinfo{journal}{Journal of Machine Learning Research}
  \bibinfo{volume}{12}, \bibinfo{pages}{3227--3257}.
\bibitem[{Ku{\ss}(2006)}]{Kuss2006}
\bibinfo{author}{Ku{\ss}, M.}, \bibinfo{year}{2006}.
\newblock \bibinfo{title}{{Gaussian Process Models for Robust Regression,
  Classification, and Reinforcement Learning}}.
\newblock Ph.D. thesis. Technische Universit\"at.
  \bibinfo{address}{{Darmstadt}}.
\bibitem[{{Li} et~al.(2020){Li}, {Shao}, {Li}, {Yu}, {Zhong} and
  {Chen}}]{Li2020a}
\bibinfo{author}{{Li}, L.}, \bibinfo{author}{{Shao}, Z.},
  \bibinfo{author}{{Li}, Z.Z.}, \bibinfo{author}{{Yu}, J.},
  \bibinfo{author}{{Zhong}, J.}, \bibinfo{author}{{Chen}, L.},
  \bibinfo{year}{2020}.
\newblock \bibinfo{title}{{Modeling Unresolved Binaries of Open Clusters in the
  Color-Magnitude Diagram. I. Method and Application of NGC 3532}}.
\newblock \bibinfo{journal}{\apj} \bibinfo{volume}{901}, \bibinfo{pages}{49}.
\newblock \DOIprefix\doi{10.3847/1538-4357/abaef3},
  \href{http://arxiv.org/abs/2008.04684}{{\tt arXiv:2008.04684}}.
\bibitem[{Liu et~al.(2020)Liu, Ong, Shen and Cai}]{Liu2020a}
\bibinfo{author}{Liu, H.}, \bibinfo{author}{Ong, Y.S.}, \bibinfo{author}{Shen,
  X.}, \bibinfo{author}{Cai, J.}, \bibinfo{year}{2020}.
\newblock \bibinfo{title}{When {{Gaussian Process Meets Big Data}}: {{A
  Review}} of {{Scalable GPs}}}.
\newblock \bibinfo{journal}{IEEE Transactions on Neural Networks and Learning
  Systems} \bibinfo{volume}{31}, \bibinfo{pages}{4405--4423}.
\newblock \DOIprefix\doi{10.1109/TNNLS.2019.2957109}.
\bibitem[{Maronna et~al.(2019)Maronna, Martin, Yohai and
  {Salibi{\'a}n-Barrera}}]{Maronna2019}
\bibinfo{author}{Maronna, R.A.}, \bibinfo{author}{Martin, R.D.},
  \bibinfo{author}{Yohai, V.J.}, \bibinfo{author}{{Salibi{\'a}n-Barrera}, M.},
  \bibinfo{year}{2019}.
\newblock \bibinfo{title}{Robust {{Statistics}}: {{Theory}} and {{Methods}}
  (with {{R}})}.
\newblock \bibinfo{publisher}{{John Wiley \& Sons}}.
\bibitem[{{Martinez-Cantin} et~al.(2018){Martinez-Cantin}, Tee and
  McCourt}]{Martinez-Cantin2018a}
\bibinfo{author}{{Martinez-Cantin}, R.}, \bibinfo{author}{Tee, K.},
  \bibinfo{author}{McCourt, M.}, \bibinfo{year}{2018}.
\newblock \bibinfo{title}{Practical {{Bayesian}} optimization in the presence
  of outliers}, in: \bibinfo{booktitle}{International {{Conference}} on
  {{Artificial Intelligence}} and {{Statistics}}}, \bibinfo{publisher}{{PMLR}}.
  pp. \bibinfo{pages}{1722--1731}.
\bibitem[{{McClintock} et~al.(2019){McClintock}, {Rozo}, {Becker}, {DeRose},
  {Mao}, {McLaughlin}, {Tinker}, {Wechsler} and {Zhai}}]{McClintock2019a}
\bibinfo{author}{{McClintock}, T.}, \bibinfo{author}{{Rozo}, E.},
  \bibinfo{author}{{Becker}, M.R.}, \bibinfo{author}{{DeRose}, J.},
  \bibinfo{author}{{Mao}, Y.Y.}, \bibinfo{author}{{McLaughlin}, S.},
  \bibinfo{author}{{Tinker}, J.L.}, \bibinfo{author}{{Wechsler}, R.H.},
  \bibinfo{author}{{Zhai}, Z.}, \bibinfo{year}{2019}.
\newblock \bibinfo{title}{{The Aemulus Project. II. Emulating the Halo Mass
  Function}}.
\newblock \bibinfo{journal}{\apj} \bibinfo{volume}{872}, \bibinfo{pages}{53}.
\newblock \DOIprefix\doi{10.3847/1538-4357/aaf568},
  \href{http://arxiv.org/abs/1804.05866}{{\tt arXiv:1804.05866}}.
\bibitem[{{Naish-Guzman} and Holden(2007)}]{Naish-Guzman2007}
\bibinfo{author}{{Naish-Guzman}, A.}, \bibinfo{author}{Holden, S.},
  \bibinfo{year}{2007}.
\newblock \bibinfo{title}{Robust regression with twinned {{Gaussian}}
  processes}, in: \bibinfo{booktitle}{Proceedings of the 20th {{International
  Conference}} on {{Neural Information Processing Systems}}},
  \bibinfo{publisher}{{Curran Associates Inc.}}, \bibinfo{address}{{Red Hook,
  NY, USA}}. pp. \bibinfo{pages}{1065--1072}.
\bibitem[{{Neal}(1997)}]{Neal1997}
\bibinfo{author}{{Neal}, R.M.}, \bibinfo{year}{1997}.
\newblock \bibinfo{title}{{Monte Carlo Implementation of Gaussian Process
  Models for Bayesian Regression and Classification}}.
\newblock \bibinfo{journal}{arXiv e-prints} ,
  \bibinfo{pages}{physics/9701026}\href{http://arxiv.org/abs/physics/9701026}{{\tt
  arXiv:physics/9701026}}.
\bibitem[{Pison et~al.(2002)Pison, Van~Aelst and Willems}]{Pison2002}
\bibinfo{author}{Pison, G.}, \bibinfo{author}{Van~Aelst, S.},
  \bibinfo{author}{Willems, G.}, \bibinfo{year}{2002}.
\newblock \bibinfo{title}{Small sample corrections for {{LTS}} and {{MCD}}}.
\newblock \bibinfo{journal}{Metrika} \bibinfo{volume}{55},
  \bibinfo{pages}{111--123}.
\newblock \DOIprefix\doi{10.1007/s001840200191}.
\bibitem[{{Ramirez-Padron} et~al.(2021){Ramirez-Padron}, Mederos and
  Gonzalez}]{Ramirez-Padron2021}
\bibinfo{author}{{Ramirez-Padron}, R.}, \bibinfo{author}{Mederos, B.},
  \bibinfo{author}{Gonzalez, A.J.}, \bibinfo{year}{2021}.
\newblock \bibinfo{title}{Robust weighted {{Gaussian}} processes}.
\newblock \bibinfo{journal}{Computational Statistics} \bibinfo{volume}{36},
  \bibinfo{pages}{347--373}.
\newblock \DOIprefix\doi{10.1007/s00180-020-01011-0}.
\bibitem[{Ranjan et~al.(2016)Ranjan, Huang and Fatehi}]{Ranjan2016}
\bibinfo{author}{Ranjan, R.}, \bibinfo{author}{Huang, B.},
  \bibinfo{author}{Fatehi, A.}, \bibinfo{year}{2016}.
\newblock \bibinfo{title}{Robust {{Gaussian}} process modeling using {{EM}}
  algorithm}.
\newblock \bibinfo{journal}{Journal of Process Control} \bibinfo{volume}{42},
  \bibinfo{pages}{125--136}.
\newblock \DOIprefix\doi{10.1016/j.jprocont.2016.04.003}.
\bibitem[{Rasmussen and Williams(2005)}]{Rasmussen2005a}
\bibinfo{author}{Rasmussen, C.E.}, \bibinfo{author}{Williams, C.K.I.},
  \bibinfo{year}{2005}.
\newblock \bibinfo{title}{Gaussian {{Processes}} for {{Machine Learning}}}.
\newblock \bibinfo{publisher}{{The MIT Press}}.
\newblock \DOIprefix\doi{10.7551/mitpress/3206.001.0001}.
\bibitem[{Ross and Dy(2013)}]{Ross2013a}
\bibinfo{author}{Ross, J.C.}, \bibinfo{author}{Dy, J.G.}, \bibinfo{year}{2013}.
\newblock \bibinfo{title}{Nonparametric mixture of {{Gaussian}} processes with
  constraints}, in: \bibinfo{booktitle}{Proceedings of the 30th {{International
  Conference}} on {{International Conference}} on {{Machine Learning}} -
  {{Volume}} 28}, \bibinfo{publisher}{{JMLR.org}}, \bibinfo{address}{{Atlanta,
  GA, USA}}. pp. \bibinfo{pages}{III--1346--III--1354}.
\bibitem[{Rousseeuw(1984)}]{Rousseeuw1984a}
\bibinfo{author}{Rousseeuw, P.J.}, \bibinfo{year}{1984}.
\newblock \bibinfo{title}{Least {{Median}} of {{Squares Regression}}}.
\newblock \bibinfo{journal}{Journal of the American Statistical Association}
  \bibinfo{volume}{79}, \bibinfo{pages}{871--880}.
\newblock \DOIprefix\doi{10.1080/01621459.1984.10477105}.
\bibitem[{Rousseeuw and Driessen(2006)}]{Rousseeuw2006}
\bibinfo{author}{Rousseeuw, P.J.}, \bibinfo{author}{Driessen, K.},
  \bibinfo{year}{2006}.
\newblock \bibinfo{title}{Computing {{LTS Regression}} for {{Large Data
  Sets}}}.
\newblock \bibinfo{journal}{Data Mining and Knowledge Discovery}
  \bibinfo{volume}{12}, \bibinfo{pages}{29--45}.
\newblock \DOIprefix\doi{10.1007/s10618-005-0024-4}.
\bibitem[{Rousseeuw and Leroy(1987)}]{Rousseeuw1987}
\bibinfo{author}{Rousseeuw, P.J.}, \bibinfo{author}{Leroy, A.M.},
  \bibinfo{year}{1987}.
\newblock \bibinfo{title}{Robust {{Regression}} and {{Outlier Detection}}}.
\newblock \bibinfo{publisher}{{John Wiley \& Sons, Inc.}},
  \bibinfo{address}{{New York, NY, USA}}.
\bibitem[{Stegle et~al.(2008)Stegle, Fallert, MacKay and Brage}]{Stegle2008}
\bibinfo{author}{Stegle, O.}, \bibinfo{author}{Fallert, S.V.},
  \bibinfo{author}{MacKay, D.J.C.}, \bibinfo{author}{Brage, S.},
  \bibinfo{year}{2008}.
\newblock \bibinfo{title}{Gaussian {{Process Robust Regression}} for {{Noisy
  Heart Rate Data}}}.
\newblock \bibinfo{journal}{IEEE Transactions on Biomedical Engineering}
  \bibinfo{volume}{55}, \bibinfo{pages}{2143--2151}.
\newblock \DOIprefix\doi{10.1109/TBME.2008.923118}.
\bibitem[{Stein(1999)}]{Stein1999}
\bibinfo{author}{Stein, M.L.}, \bibinfo{year}{1999}.
\newblock \bibinfo{title}{Interpolation of {{Spatial Data}}: {{Some Theory}}
  for {{Kriging}}}.
\newblock Springer {{Series}} in {{Statistics}},
  \bibinfo{publisher}{{Springer-Verlag}}, \bibinfo{address}{{New York}}.
\newblock \DOIprefix\doi{10.1007/978-1-4612-1494-6}.
\bibitem[{Vanhatalo et~al.(2009)Vanhatalo, Jyl{\"a}nki and
  Vehtari}]{Vanhatalo2009}
\bibinfo{author}{Vanhatalo, J.}, \bibinfo{author}{Jyl{\"a}nki, P.},
  \bibinfo{author}{Vehtari, A.}, \bibinfo{year}{2009}.
\newblock \bibinfo{title}{Gaussian process regression with {{Student}}-t
  likelihood}, in: \bibinfo{editor}{Bengio, Y.}, \bibinfo{editor}{Schuurmans,
  D.}, \bibinfo{editor}{Lafferty, J.D.}, \bibinfo{editor}{Williams, C.K.I.},
  \bibinfo{editor}{Culotta, A.} (Eds.), \bibinfo{booktitle}{Advances in
  {{Neural Information Processing Systems}} 22}. \bibinfo{publisher}{{Curran
  Associates, Inc.}}, pp. \bibinfo{pages}{1910--1918}.
\bibitem[{Wang et~al.(2017)Wang, Xue, Cui and Zhong}]{Wang2017c}
\bibinfo{author}{Wang, D.}, \bibinfo{author}{Xue, J.}, \bibinfo{author}{Cui,
  D.}, \bibinfo{author}{Zhong, Y.}, \bibinfo{year}{2017}.
\newblock \bibinfo{title}{A robust submap-based road shape estimation via
  iterative {{Gaussian}} process regression}, in: \bibinfo{booktitle}{2017
  {{IEEE Intelligent Vehicles Symposium}} ({{IV}})},
  \bibinfo{publisher}{{IEEE}}, \bibinfo{address}{{Los Angeles, CA, USA}}. pp.
  \bibinfo{pages}{1776--1781}.
\newblock \DOIprefix\doi{10.1109/IVS.2017.7995964}.
\bibitem[{Yohai(1987)}]{Yohai1987}
\bibinfo{author}{Yohai, V.J.}, \bibinfo{year}{1987}.
\newblock \bibinfo{title}{High {{Breakdown}}-{{Point}} and {{High Efficiency
  Robust Estimates}} for {{Regression}}}.
\newblock \bibinfo{journal}{The Annals of Statistics} \bibinfo{volume}{15},
  \bibinfo{pages}{642--656}.
\newblock \DOIprefix\doi{10.1214/aos/1176350366}.
\bibitem[{Yu and Yao(2017)}]{Yu2017}
\bibinfo{author}{Yu, C.}, \bibinfo{author}{Yao, W.}, \bibinfo{year}{2017}.
\newblock \bibinfo{title}{Robust linear regression: {{A}} review and
  comparison}.
\newblock \bibinfo{journal}{Communications in Statistics - Simulation and
  Computation} \bibinfo{volume}{46}, \bibinfo{pages}{6261--6282}.
\newblock \DOIprefix\doi{10.1080/03610918.2016.1202271}.

\end{thebibliography}

\end{document}